\documentclass[twoside,a4paper]{article}

\usepackage{arxiv}

\usepackage[utf8]{inputenc} 
\usepackage[T1]{fontenc}    
\usepackage{times}
\usepackage{epsfig}
\usepackage{graphicx}
\usepackage{amsmath}
\usepackage{amssymb}
\usepackage{longtable}
\usepackage{makecell}
\usepackage{ulem}
\usepackage{mathrsfs}
\usepackage{cite}
\usepackage{graphicx}
\usepackage{multirow}
\usepackage{tabularx}
\usepackage[table]{xcolor}    
\usepackage{float}

\usepackage[pagebackref=true,breaklinks=true,bookmarks=false]{hyperref}

\def\cvwwPaperID{****} 

\begin{document}

\title{BIRL: Benchmark on Image Registration methods {\small with}~Landmark validation}

\author{Jiri Borovec\\
FEE, Czech Technical University in Prague\\
{\tt\small jiri.borovec@fel.cvut.cz}
}

\maketitle

\begin{abstract}
   This report presents a generic image registration benchmark with automatic evaluation using landmark annotations. 
   The key features of the BIRL framework are: 
   easily extendable,
   performance evaluation,
   parallel experimentation,
   simple visualisations,
   experiment's time-out limit,
   resuming unfinished experiments.
   From the research practice, we identified and focused on these two main use-cases:
    (a) comparison of user's (newly developed) method with some State-of-the-Art (SOTA) methods on a common dataset and
    (b) experimenting SOTA methods on user's custom dataset (which should contain landmark annotation). 
   
   Moreover, we present an integration of several standard image registration methods aiming at biomedical imaging into the BIRL framework. 
   This report also contains experimental results of these SOTA methods on the CIMA dataset, which is a dataset of Whole Slice Imaging (WSI) from histology/pathology containing several multi-stain tissue samples from three tissue kinds.
   
   {Source and results: \tt\small \url{https://borda.github.io/BIRL}}
\end{abstract}



\newcommand{\vc}[1]{{\boldsymbol{\mathbf #1}}}
\newcommand{\lnd}{{\vc{x}}} 
\newcommand{\lndW}{\hat{\lnd}} 
\newcommand{\Lnd}{L} 
\newcommand{\LndS}{K} 
\newcommand{\fix}{F}
\newcommand{\mov}{M}
\newcommand{\wrp}{W}


\section{Introduction}

The image registration is a crucial task in several domains, although this report focuses mainly on biomedical image registration~\cite{Maintz1998, zitova2003, Salvi2007, Sotiras2013, Alam2017, Haskins2019, Alam2019} and in particular Whole Slice Imaging (WSI) in histology/pathology~\cite{Deniz2015}, but it can be easily reused also in other domains such as material analyses,  surveillance, etc.

In digital pathology, ~\cite{Gurcan2009}, one of the most simple and yet most useful features is the ability to view serial sections of tissue simultaneously on a computer screen~\cite{West1997, Murphy2011, Metzger2012, Garcia2010, Novakovic2012}.
This enables the pathologist to evaluate the histology and expression of multiple markers for a patient in a single review~\cite{Thul2018}.
However, the rate-limiting step in this process is the time taken for the pathologist to open each individual image, align the sections within the viewer, and then manually move around the section.
In addition, due to tissue processing and pre-analytical steps, sections with different stains have non-linear variations among their acquisitions; also they may stretch and change shape from section to section, or some tissue fraction may be damaged or missing.~\cite{Chin2007, Lopez2014, Song2013, Kartasalo2016, Kartasalo2018}

It is generally known that the WSI image registration is not a well-solved problem compare to other single-modal tasks like MRI, CT or ultrasound.
The multi-stain WSI registration can be assumed to be an almost multi-modal problem since the variety of used stains dramatically change appearance model, and the deformation may range from fine elastic transformation to completely missing section (due to mechanical processes~/~sample preparations) and flopping samples while image sensing.~\cite{Borovec2013, Kybic2014, Deniz2015, Kartasalo2016}
In recent years we notice a steady number of papers aiming at WSI image registration, but quite often they miss fair comparison of their newly developed method with well-established methods for the particular domain.~\cite{Borovec2018}
This became a primary motivation to collect and annotated a histology dataset of WSI microscopy images and develop an image registration evaluation framework to fill this gap.

In the past, there was a few image registration benchmarks~\cite{West1997, Castillo2009, Ou-TMI2014, Christensen2006, Klein2009} and challenges~\cite{Murphy2011, Marstal2019}.
Typically, they were focusing on a very narrow domain (single-modal) and using relatively small images (a thousand pixels compare to typical WSI with tens of thousands of pixels in image diagonal).
The limitation of the last attend~\cite{Marstal2019} to democratise image registration benchmarking is the need of tight integration to their framework (which may lead even to a time-costly rewriting whole method) compare to our BIRL which can run with almost anything.


Let us briefly talk about the context and history of this work. 
First, we introduced a benchmark~\cite{Borovec2018} comparing several methods on quite small images up to 5k pixels (with respect to WSI sizes) where not all images were allowed to be public.
Later the BIRL framework was redesigned, and we introduced an ANHIR\footnote{\url{https://anhir.grand-challenge.org}} also challenge with some hidden/private data for method's evaluation.
Finally, we present only publicly available image registration methods experimented on a publicly available CIMA dataset.


\section{BIRL framework}

The BIRL is a light-weighted Python framework for easy image registration experimentation and benchmarking on landmarks-like annotated datasets.

\begin{figure}[th]
  \centering
    \includegraphics[width=0.95\textwidth]{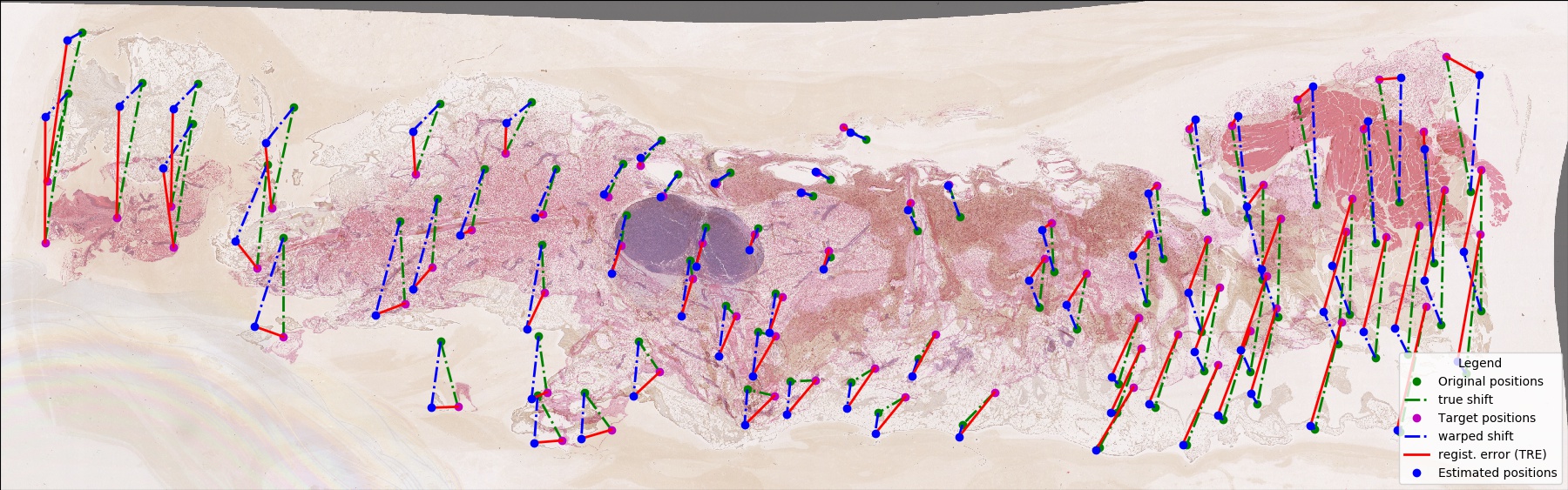}
    \\
    \includegraphics[width=0.95\textwidth]{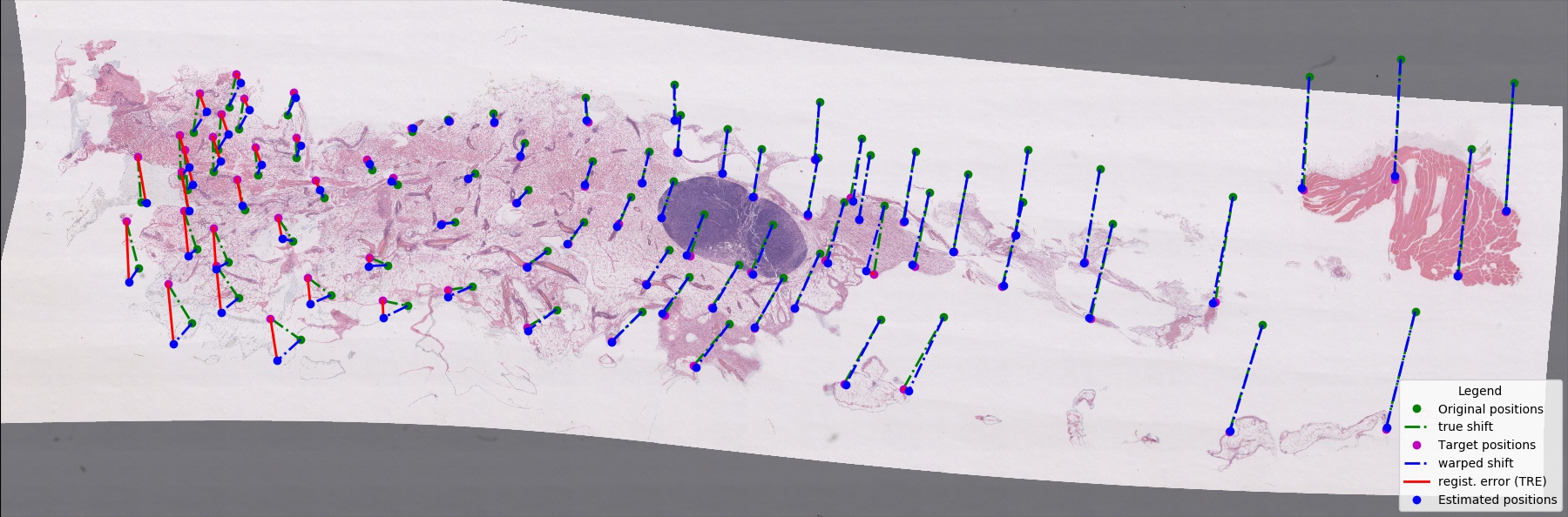}
    \\
    \includegraphics[width=0.95\textwidth]{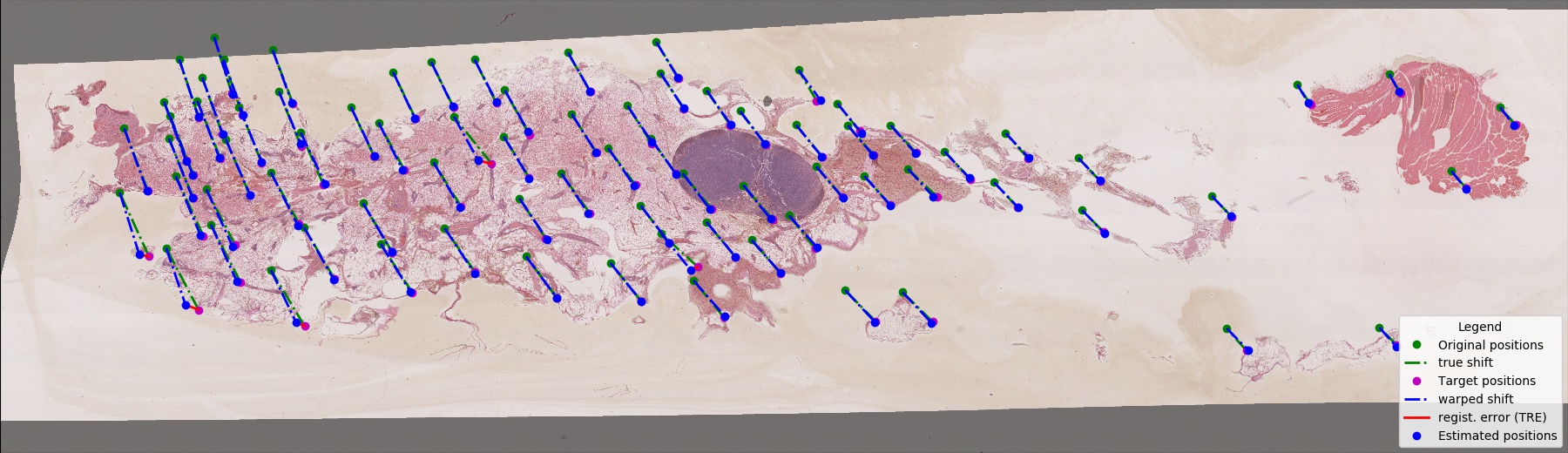}
  \caption{\label{fig:image-tre}Sample visualisation of TRE for wrongly (top), partially well (middle) and fine (bottom) registered image pairs.
  The lines represent the relation between particular landmarks in moving/fixed/warped images: (green) the true mapping, (blue) estimated mapping and (red) TRE.}
\end{figure}

Let us summarise the main/key features of this framework:
\begin{itemize}
    \item automatic executing image registration on a sequence of image pairs
    \item integrated evaluation of registration performances using relative Target Registration Error (rTRE)
    \item integrated visualisations of performed registrations
    \item running several image registration experiments in parallel
    \item resuming unfinished sequence of registration benchmark
    \item creating/handling custom dataset and design its own experiments
    \item utilising basic image pre-processing, e.g. colour normalising
    \item stand-alone post-processing: evaluation and visualisation for finished benchmarks
\end{itemize}

Moreover, the framework is developed as an installable package so any particular functionality can be reused in other similar projects, for example calling the same metrics or handling datasets.

\subsection{Benchmark workflow}

Then core benchmark \textit{class} is designed to be inherited and just needed experimental calls to be overwritten in a (\textit{child}) class for a specific image registration method.
In particular, the benchmark workflow is the following:
\begin{enumerate}
    \item preparing the experiment's environment, e.g. create experiment folder, copy configurations, etc.
    \item loading required data - parsing the experiment's image/landmarks pairs;
    \item performing the benchmark sequence (optionally in parallel) and save particular results (registration outputs and statistic) to a common table and optionally create a visualisation of performed experiments;
    \item evaluating results overall performed experiments;
    \item summarising and exporting results from the complete benchmark.
\end{enumerate}

\subsection{Metrics}

There are a few evaluation approaches which are derived according to available datasets' annotations.
The most common and less time demanding to obtain adequate annotation is using landmarks and measure their distances~\cite{West1997, Castillo2009, Brock-IJRO2010, Ou-TMI2014, Borovec2018}, another option is using image segmentation and measure its overlap~\cite{Christensen2006, Klein2009, Murphy2011}.
The precision of both measures is sensitive on the landmarks density or details in segmentation (meaning the number of utilised classes and degree of detail captured) respectively.

As it was stated in the title, we use landmark-based evaluation.
Assume we have have set of landmarks (key points marking uniquely identifiable structures) $\Lnd^\fix$ and $\Lnd^\mov$ in fixed $I^\fix$ and moving $I^\mov$ image respectively and particular landmarks $\lnd^\fix \in \Lnd^\fix$ and $\lnd^\mov \in \Lnd^\mov$ marking the same biological structure in both images.
Moreover, we have a set of warped landmarks $\lndW^\mov$ from $\lnd^\mov$ to match $\lnd^\fix$.

The evaluation is based on Target Registration Error (TRE) between two sets of landmarks $\lnd^\fix$ and $\lndW^\mov$ in the two images.
Lets us denote
\begin{equation}
    TRE=d_e(\lnd_l^\fix,\lndW_l^\mov)
\end{equation}
where $\lnd^\fix$ and $\lndW^\mov$ are the associated landmark coordinates and $d_e(.)$ is the Euclidean distance. 
As image sizes across datasets differ making an overall comparison uneasy, all TRE are normalised by the image diagonal of the fixed image $I^\fix$ to a relative TRE (rTRE), 
\begin{equation}
    rTRE=\frac{TRE}{\text{diagonal($I^\fix$)}}
\end{equation}

As a successful-rate-like measure, we introduce robustness $R$ as a relative value describing how many landmarks $\Lnd$ improved its TRE by performed image registration compared to the initial TRE, otherwise, formally 
\begin{equation}
    R = \cfrac{1}{|\Lnd|} \sum_{l \in \Lnd} [\![ TRE(\lnd_l^\fix,\lndW_l^\mov) < TRE(\lnd_l^\fix,\lnd_l^\mov) ]\!]
\end{equation}
All failing image registrations are considered to have the initial position $\lndW^\mov \xleftarrow{} \lnd^\mov$.

Let us also denote following aggregation measures over single experiment (registration image pair): $m_i(.)=\text{median}_\text{image}(.)$ and $s_i(.)=\text{max}_\text{image}(.)$; and over the dataset $a_d(.)=\text{mean}_\text{dataset}(.)$, $m_d(.)=\text{median}_\text{dataset}(.)$, which are used in a cascade - meaning first compute measures over particular images then aggregated over whole dataset, for example $a_d(m_i(rTRE(\lnd_l^\fix,\lndW_l^\mov)))$ which is later shorten to $AMrTRE$.
The motivation for using the median is a lower penalisation for a few inaccurate landmarks while most of them are registered well.
These statistics also assumes a uniform distribution of landmarks annotation over examining tissue sample.

\subsection{Typical use-cases}

We have identified the two most common use-cases for this framework - how a user can most benefits from this work with minimal effort.

\paragraph{Comparing with SOTA on a common dataset.}

The quite common problem with newly developed methods is presenting only their results on a private custom dataset, which may have a pure description, moreover, usually the dataset is small.
This missing comparison with SOTA methods on the standard (well described) dataset can be fixed by integrating the new method to the BIRL framework, run the benchmark and compare its new results with the presented scores.
For this case user need to overwrite an only fraction of methods/functions which is essential for their image registration method and its evaluation:
\begin{itemize}
    \item \textit{\_prepare\_img\_registration(...)} using if some extra preparation before running own image registration is needed, e.g. converting images to different format. [before each image registration experiment]
    \item \textit{\_execute\_img\_registration} executing/performing the image registration, time of this method is measured as execution time; in case user calls external method from a command line, he rewrites only \textit{\_generate\_regist\_command(...)} which prepares the registration command to be executed automatically. [core of each image registration experiment]
    \item \textit{\_extract\_warped\_image\_landmarks(...)} extracting the required warped landmarks or perform landmark warping in this stage if it was not already done as a part of the image registration. [after each image registration experiment]
    \item \textit{\_clear\_after\_registration(...)} removing some temporary files generated during image registration to keep the results lightweight. [after each image registration experiment]
\end{itemize}

\paragraph{Exploring SOTA methods on a custom dataset.} 

This is another practical use-case, especially for biomedical experts starting with a discovery on a new dataset. 
We suppose the user may ask a question: "What is the best method for aligning my new images together?"
Then he can prepare dataset pairing - CSV table with rows containing paths to the source and target image and their landmarks annotation.
If a performance evaluation/analyses are not required, and a simple visual evaluation is sufficient, the landmark annotation can be omitted.


\section{Experiments}

With the framework in our hands, we added some State-Of-The-Art (SOTA) methods and ran them on a prepared dataset.
We start with a brief description of integrated SOTA methods, followed by recapitulation on CIMA dataset and finishing with presenting the results of the SOTA methods on this WSI dataset (also showing visualisations produced by this framework).

\subsection{Standard image registration methods}

There are many standards and/or widely used methods/software/frameworks for biomedical image registration.
We stick only with publicly available ones and those who are capable of producing warped landmarks based on estimated transformation for our performance evaluation.

\paragraph{Advanced Normalisation Tools (ANTs)~\cite{Avants2008}} is a registration toolkit using Insight Segmentation and Registration Toolkit (ITK)\footnote{\url{https://itk.org}} as a~backend aiming at MR imaging.
The ANTs allows creating custom image registration pipeline composed of several transformations and similarity measures in a multi-scale scheme.
[\textit{In our experiments, we used a combination of an~affine registration with Mattes Mutual Information (MMI) followed by SyN registration with Cross-Correlation (CC) similarity measure.}]

\paragraph{bUnwarpJ~\cite{Arganda2006}} is a ImageJ/Fiji~\cite{FIJI2012} plugin which estimates a~symetric non-linear B-spline-based deformation.
The minimised criterion is a~sum of squares difference (SSD) in a multi-resolution scheme. [\textit{We have experimented two versions: (bUnwarpJ) using only the image intensities and (bUnwarpJ+SIFT) combining image intensity with SIFT detector equally balancing both fractions.}]

\paragraph{DROP~\cite{Glocker2008, Glocker2011}} differs from most other methods by using discreet optimisation (solving efficiently in a~multiresolution
fashion using linear programming) for minimising a~sum of absolute differences (SAD) criterion.

\paragraph{Elastix~\cite{Klein2010}} is an image registration software base on the ITK offering several transformations/metrics/optimisations in multi-resolution scheme.
[\textit{In our experiments, we used b-spline image registration with Adaptive Stochastic Gradient Descent minimising Advanced MMI criterion.}]

\paragraph{NiftyReg~\cite{Ourselin2001,Modat2014}} is an open-source software performing linear and/or nonlinear registration for two and three dimensional images. 
The linear registration is based on a block-matching technique, and the non-linear is using the Free-Form Deformation.
[\textit{In our experiments, we used R wrapper to this software and defined two-step registration - starting with linear and followed by non-linear.}]

\paragraph{Register Virtual Stack Slices (RVSS)~\cite{Arganda2006}} is another ImageJ/Fiji registration plugin which extends the bUnwarpJ and relies on SIFT~\cite{lowe04} feature points offering several deformation types.
[\textit{In our experiments, we used similarity for feature matching and affine for the final alignment.}]

Note that methods allowing other configuration of transformation, metric, optimisation, etc. can be easily reconfigured within this benchmark (with a simple parameter file) without any code changes.


\subsection{CIMA dataset}

\begin{table}[t]
  \begin{center}
    \begin{tabular}{c|l}
    \textbf{Code} & \textbf{Name}                   \\
    \hline
    Cc10 &      Clara cell 10 protein \\
    CD31 &      Platelet endothelial cell adhesion molecule \\
    ER &        Estrogen receptor \\
    H\&E &      Hematoxylin and Eosin \\
    HER-2 &     Human epidermal growth factor receptor 2 \\
    Ki67 &      Antigen KI-67 \\
    PR &        Progesterone receptor \\
    proSPC &    Prosurfactant protein C \\
    \end{tabular}
    \caption{\label{tab:stains}
    Complete list of biological markers used for staining histology tissue samples.
    }
    
  \end{center}
\end{table}

The dataset\footnote{\url{http://cmp.felk.cvut.cz/~borovji3/?page=dataset}}~\cite{Fernandez2002} consists WSI of microscopy images of 2D pathological tissue slices, stained with different markers (for stain explanation see Tab.~\ref{tab:stains}), and landmarks (manually annotated uniquely unidentified biological structure) in each slice. 
The main challenges for these images are the following: (i) enormous image size, (ii) appearance differences, and (iii) lack of distinctive appearance objects. 
In particular, this dataset contains nine tissue samples of three different tissue types which together form 108 image pairs.

Let us start with a short description of the particular tissue samples, followed by landmarks and forming registration image pairs.
For the dataset overview, see Tab.~\ref{tab:dataset}.

Brief description of the three tissue types included in the CIMA datasets:

\begin{itemize}
    \item \textbf{Lung lesion}
Unstained adjacent $3 \mu \hbox{m}$ formalin-fixed paraffin-embedded sections were cut from the blocks and stained with H\&E or by immunohistochemistry with a specific antibody for CD31, proSPC, CC10 or Ki67.
Images of three mice lung lesions (adenoma or adenocarcinoma) were acquired with a Zeiss Axio Imager M1 microscope (Carl Zeiss, Jena, Germany) equipped with a dry Plan Apochromat objective (numerical aperture $\text{NA}=0.95$, magnification $40\times$, pixel size $0.174 \mu \hbox{m}/\text{pixel}$).
    \item \textbf{Lung lobes}
The images of the four whole mice lung lobes correspond to the same set of histological samples as the lesion tissue.
They were also acquired with a Zeiss Axio Imager M1 microscope (Carl Zeiss, Jena, Germany) equipped with a dry EC Plan-Neofluar objective ($\text{NA}=0.30$, magnification $10\times$, pixel size $1.274 \mu \hbox{m}/\text{pixel}$).
    \item \textbf{Mammary glands}
The sections are cuts from two mammary glands blocks stained with H\&E (even sections) and alternatively, with an antibody against the ER, PR, or Her2-neu (odd sections).
They were also acquired with a Zeiss Axio Imager M1 microscope (Carl Zeiss, Jena, Germany) equipped with a dry EC Plan-Neofluar objective ($\text{NA}=0.30$, magnification $10\times$, pixel size $1.274 \mu \hbox{m}/\text{pixel}$).
\end{itemize}

\begin{table}[t]
  \begin{center}
    \begin{tabular}{l|ccccc}

Name    &    \thead{$\mu \hbox{m}/$pixel}  &    \thead{images \\ in set}    &    \thead{points \\ per image}    &    \thead{Avg. size\\ $[\text{pixels}]$}    &    \thead{10k zoom \\ $[\%]$ } \\
\hline
Lung Lesions 1    &   $0.174$       &    5    &    78    &    $16k$ & 50 \\
Lung Lesions 2    &   $0.174$       &    5    &    101    &    $23k$ & 25 \\
Lung Lesions 3    &   $0.174$       &    5    &    80    &    $16k$ & 50 \\
Lung lobes 1    &   $1.274$       &    5    &    98    &    $10k$ & 100 \\
Lung lobes 2    &   $1.274$       &    5    &    107    &    $10k$ & 100 \\
Lung lobes 3    &   $1.274$       &    5    &    80    &    $9k$ & 100 \\
Lung lobes 4    &   $1.274$       &    5    &    86    &    $9k$ & 100 \\
Mammary gland 1    &   $2.294$       &    5    &    82    &    $22k$ & 25 \\ 
Mammary gland 2    &   $2.294$       &    8    &    76    &    $20k$ & 25 \\
    \end{tabular}
    \caption{\label{tab:dataset}
    Dataset summary per tissue sample with used scales for $10k$ dataset scale.}
    
  \end{center}
\end{table}

\begin{table}[t]
  \begin{center}
    \begin{tabular}{l|ccccc}
             &   Cc10    &    CD31    &    H\&E    &   Ki67    &    proSPC  \\
     \hline
    Cc10    &   -       &    1       &    2       &    3       &    4       \\
    CD31    & \sout{1}  &    -       &    5       &    6       &    7       \\
    H\&E    & \sout{2}  & \sout{5}  &    -       &    8       &    9       \\
    Ki67    & \sout{3}  & \sout{6}  & \sout{8}  &    -       &    10      \\
    proSPC    & \sout{4}  & \sout{7}  & \sout{9}  & \sout{10} &    -       \\
    \end{tabular}
    \caption{\label{tab:pairing}An example of forming 10 registration pairs from a set of 5 differently stained images.
    The equal pairs indexes in upper-right and lower-left parts mark the mirroring pairs where the strike ones are omitted.}
    
  \end{center}
\end{table}

\paragraph{Landmarks}
The landmarks as mentioned earlier are points marking unique biological structures appearing in all images (tissue slice) of a single tissue sample where each slice is stain by a different marker.
The landmarks for each image (stain tissue slice) are stored in a table (a CSV file) where the lines match the same biological structures across stain slices.
This format is very intuitive, and it has a standard ImageJ structure, and coordinate frame - the origin $(0, 0)$ of the coordinate system is set to the image top left corner.
On the side of annotation quality, each landmark annotation was performed by two experts, and each annotation was validated by another expert independently.
Further information about landmarks and annotation procedure is available on web\footnote{\url{https://borda.github.io/dataset-histology-landmarks}} along with annotation tools which allow to add new landmarks and share them among all dataset users.

\paragraph{Pairing images}
As the stain slices within each tissue sample are very close each to other and we have unified annotations over all slices in a set, we register all image (slices) to each other which increase the number of registration pairs, see Tab.~\ref{tab:pairing}.
We assume that the registration of two images (fixed and moving) is symmetric $I^\mov \xrightarrow{} I^\fix$ and $I^\mov \xleftarrow{} I^\fix$, so we drop mirroring pairs.

\subsection{Experimental setting}

We run experiments on Linux server with 24 CPU and 250GB RAM.
As its performance is above standard machine, the framework uses an option to normalise execution time to any other reference (define) machine.
Moreover, we can run four experiments in parallel.

The dataset allows using scales.
We experiment on WSI (denoted as "$full$") and also a mix size images close to $10k$ pixels in image diagonal (denoted as "$10k$"), the same as ANHIR challenge did (to see particular scales used in $10k$ see Tab.~\ref{tab:dataset}).

We set hard time-out limit $3$ hours per single image registration; if a method does not finish in the frame time, it is terminated and considered as a failure case.

The best-parameters for each used method was setup based on authors recommendations (e.g. bUnwarpJ, RVSS and DROP) or experimentally if the first option was not feasible.\footnote{All the used configuration can be found on the BIRL project-page.}

\subsection{Results}

\begin{table*}[!t]
  \begin{center}
      
\definecolor{lightgray}{gray}{0.9}


\begin{tabular}{|l|@{\hskip 2pt}c@{\hskip 2pt}||r@{\hskip 6pt}c|c|r@{\hskip 6pt}c|r@{\hskip 6pt}c|c|r@{\hskip 6pt}c|}
\hline 
\multirow{2}{*}{methods} & \multirow{2}{*}{scope} & \multicolumn{3}{c|}{Median rTRE {[}\%{]}} & \multicolumn{2}{c|}{Max rTRE {[}\%{]}} & \multicolumn{3}{c|}{Robustness {[}\%{]}} & \multicolumn{2}{c|}{time {[}min{]}}\tabularnewline
\cline{3-12} 
 &  & \multicolumn{2}{c|}{Avg.~$\pm$~STD} & Median & \multicolumn{2}{c|}{Avg.~$\pm$~STD} & \multicolumn{2}{c|}{Avg.~$\pm$~STD} & Median & \multicolumn{2}{c|}{Average~$\pm$~STD}\tabularnewline
\hline 
\hline 
 & \textit{\small{}10k} & 2.30 & $\pm$2.00 & 1.67 & 5.56 & $\pm$3.66 & 79.02 & $\pm$24.82 & 89.25 & 52.17 & $\pm$26.89\tabularnewline
\cline{2-12} 
\rowcolor{lightgray}
\multirow{-2}{*\cellcolor{white}}{ANTs} & \textit{\small{}full} & 3.85 & $\pm$2.88 & 3.47 & 7.36 & $\pm$4.34 & 30.35 & $\pm$41.13 & 0 & 210.46 & $\pm$106.41\tabularnewline
\hline 
 & \textit{\small{}10k} & 2.82 & $\pm$2.08 & 3.00 & 6.66 & $\pm$3.66 & 74.07 & $\pm$27.75 & 83.13 & 2.99 & $\pm$1.13\tabularnewline
\cline{2-12} 
\rowcolor{lightgray}
\multirow{-2}{*\cellcolor{white}}{bUnwarpJ} & \textit{\small{}full} & 3.26 & $\pm$1.93 & 3.36 & 7.11 & $\pm$3.49 & 67.61 & $\pm$28.93 & 68.81 & 14.01 & $\pm$15.34\tabularnewline
\hline 
bUnwarpJ & \textit{\small{}10k} & 7.43 & $\pm$7.93 & 4.23 & 13.53 & $\pm$12.24 & 49.7 & $\pm$38.36 & 54.01 & 2.66 & $\pm$1.17\tabularnewline
\cline{2-12} 
\rowcolor{lightgray}
~~~~~+ SIFT \cellcolor{white} & \textit{\small{}full} & 5.90 & $\pm$5.36 & 4.26 & 12.40 & $\pm$12.66 & 50.97 & $\pm$34.74 & 54.01 & 11.36 & $\pm$11.59\tabularnewline
\hline 
 & \textit{\small{}10k} & 2.50 & $\pm$5.11 & 0.51 & 6.29 & $\pm$7.92 & 84.25 & $\pm$30.46 & 98.68 & 1.86 & $\pm$0.84\tabularnewline
\cline{2-12} 
\rowcolor{lightgray}
\multirow{-2}{*\cellcolor{white}}{DROP} & \textit{\small{}full} & 2.81 & $\pm$3.89 & 0.89 & 6.66 & $\pm$6.20 & 61.23 & $\pm$43.46 & 86.51 & 11.71 & $\pm$8.21\tabularnewline
\hline 
 & \textit{\small{}10k} & 3.79 & $\pm$2.90 & 3.47 & 8.29 & $\pm$4.47 & 79.55 & $\pm$16.48 & 80.77 & 4.02 & $\pm$0.75\tabularnewline
\cline{2-12} 
\rowcolor{lightgray}
\multirow{-2}{*\cellcolor{white}}{Elastix} & \textit{\small{}full} & 4.11 & $\pm$2.76 & 3.80 & 8.39 & $\pm$4.37 & 70.61 & $\pm$16.10 & 69.74 & 16.92 & $\pm$11.69\tabularnewline
\hline 
 & \textit{\small{}10k} & 3.20 & $\pm$1.93 & 3.22 & 6.85 & $\pm$3.49 & 68.05 & $\pm$29.69 & 70.2 & 0.36 & $\pm$0.6\tabularnewline
\cline{2-12} 
\rowcolor{lightgray}
\multirow{-2}{*\cellcolor{white}}{RNiftyReg} & \textit{\small{}full} & 3.16 & $\pm$1.94 & 3.27 & 6.83 & $\pm$3.61 & 69.77 & $\pm$29.22 & 78.14 & 0.39 & $\pm$0.56\tabularnewline
\hline 
 & \textit{\small{}10k} & 5.95 & $\pm$15.83 & 3.24 & 9.48 & $\pm$16.54 & 57.45 & $\pm$28.93 & 54.12 & 1.5 & $\pm$0.83\tabularnewline
\cline{2-12} 
\rowcolor{lightgray}
\multirow{-2}{*\cellcolor{white}}{RVSS} & \textit{\small{}full} & 4.03 & $\pm$2.67 & 3.61 & 7.63 & $\pm$4.45 & 55.13 & $\pm$22.08 & 52.42 & 4.59 & $\pm$3.61\tabularnewline
\hline 
\end{tabular}

    \caption{\label{tab:results}Aggregated results of all standard image registration methods over both dataset scopes (sizes - $10k$ pixels in image diagonal and $full$ WSI microscopy images).}
    
  \end{center}
\end{table*}

\begin{figure}[t]
  \centering
    \includegraphics[width=0.95\textwidth]{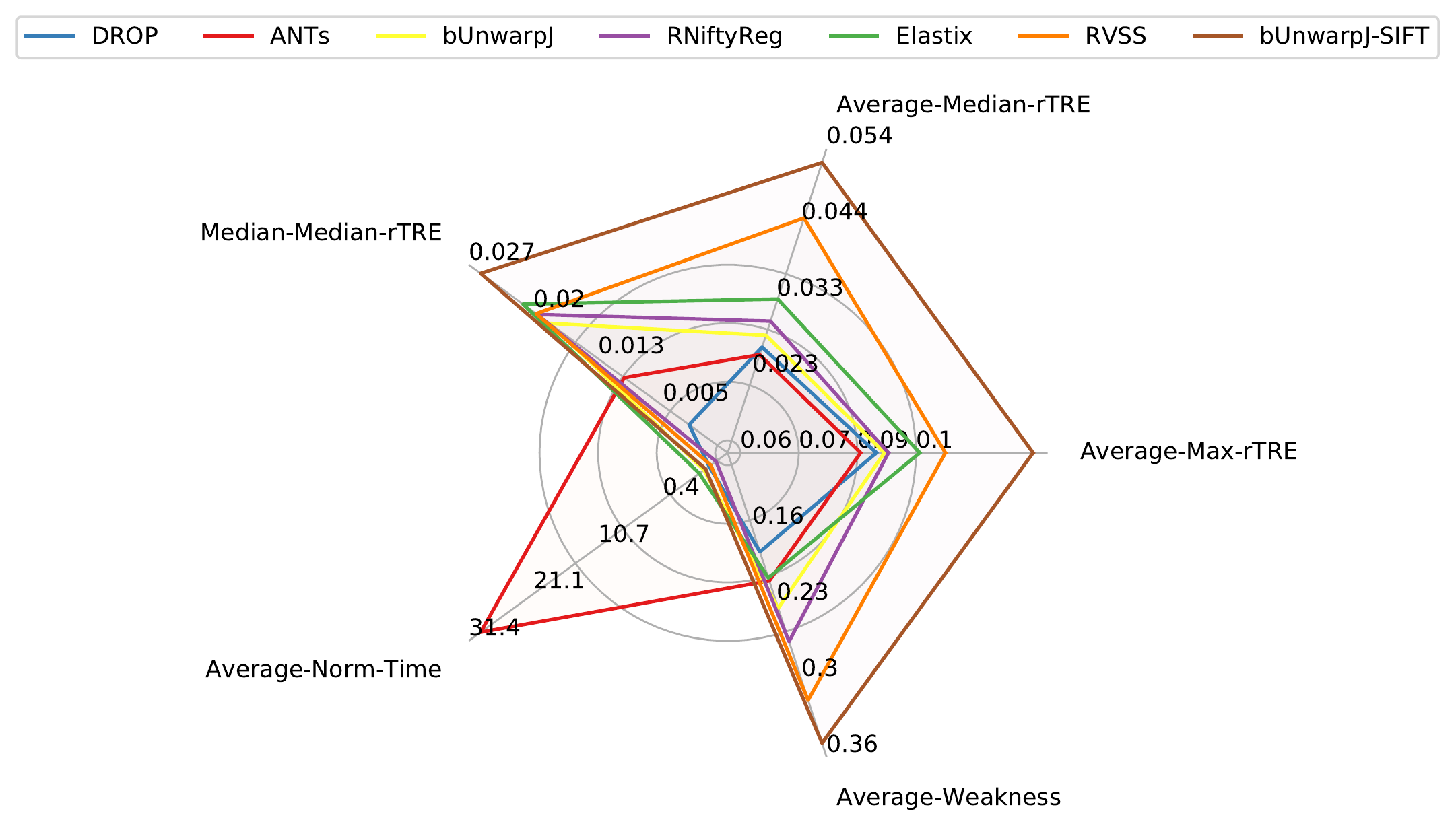}
  \caption{\label{fig:radar}Summary visualisation of benchmark results over several key metrics for all method on $10k$ dataset scope.
  For all metrics, closer to the centre, better.
  The weakness measure is inverse robustness.}
\end{figure}

\begin{figure}[t]
  \centering
    \includegraphics[width=0.49\textwidth]{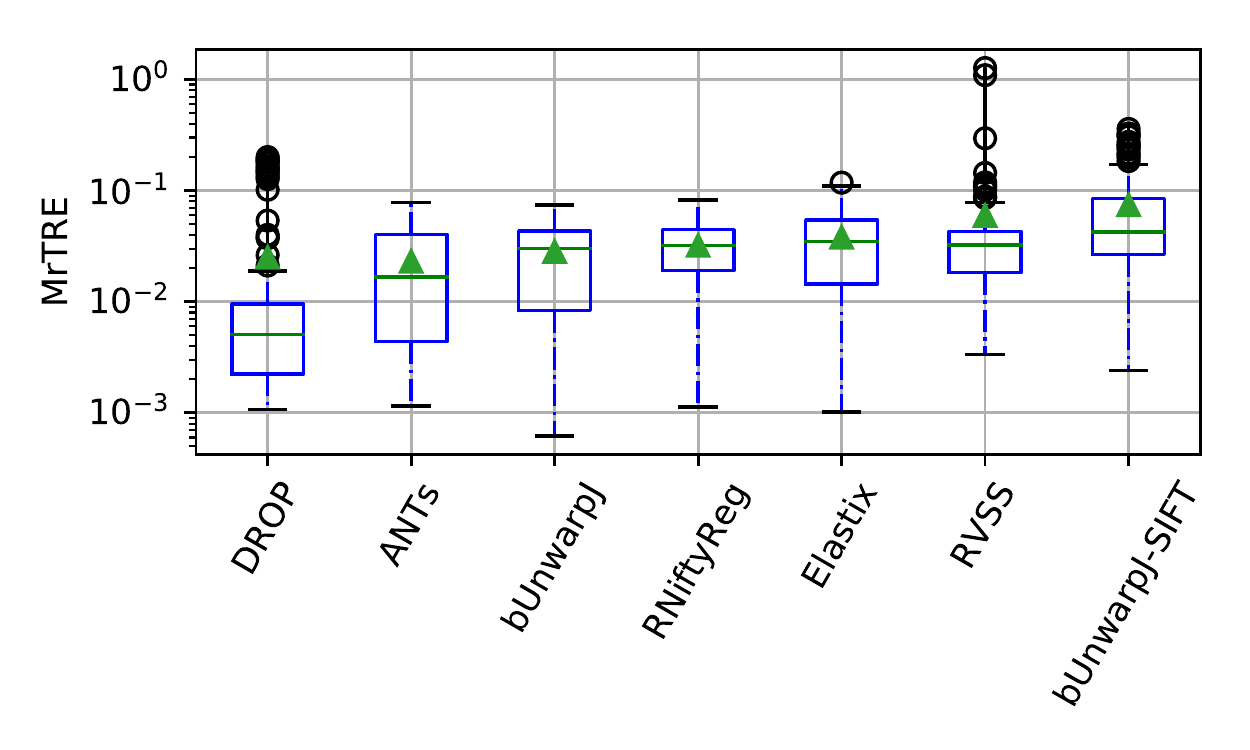}
    \includegraphics[width=0.49\textwidth]{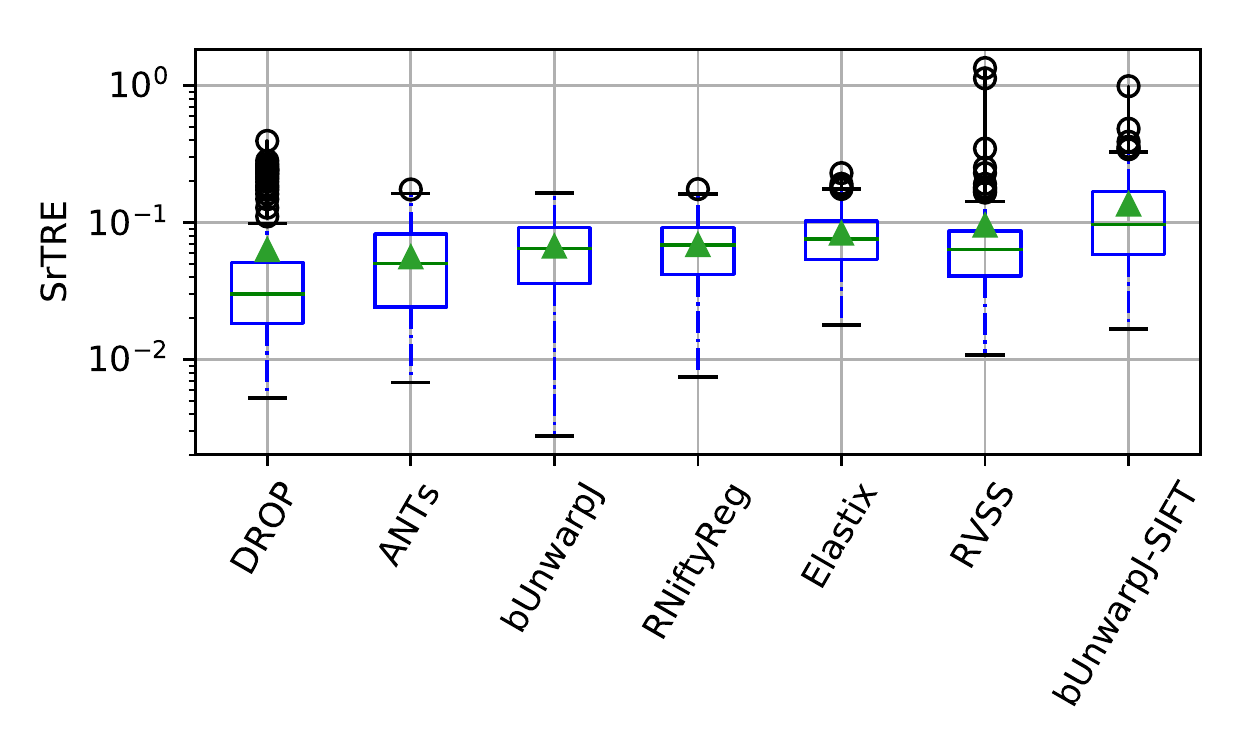}
    \\
    \includegraphics[width=0.49\textwidth]{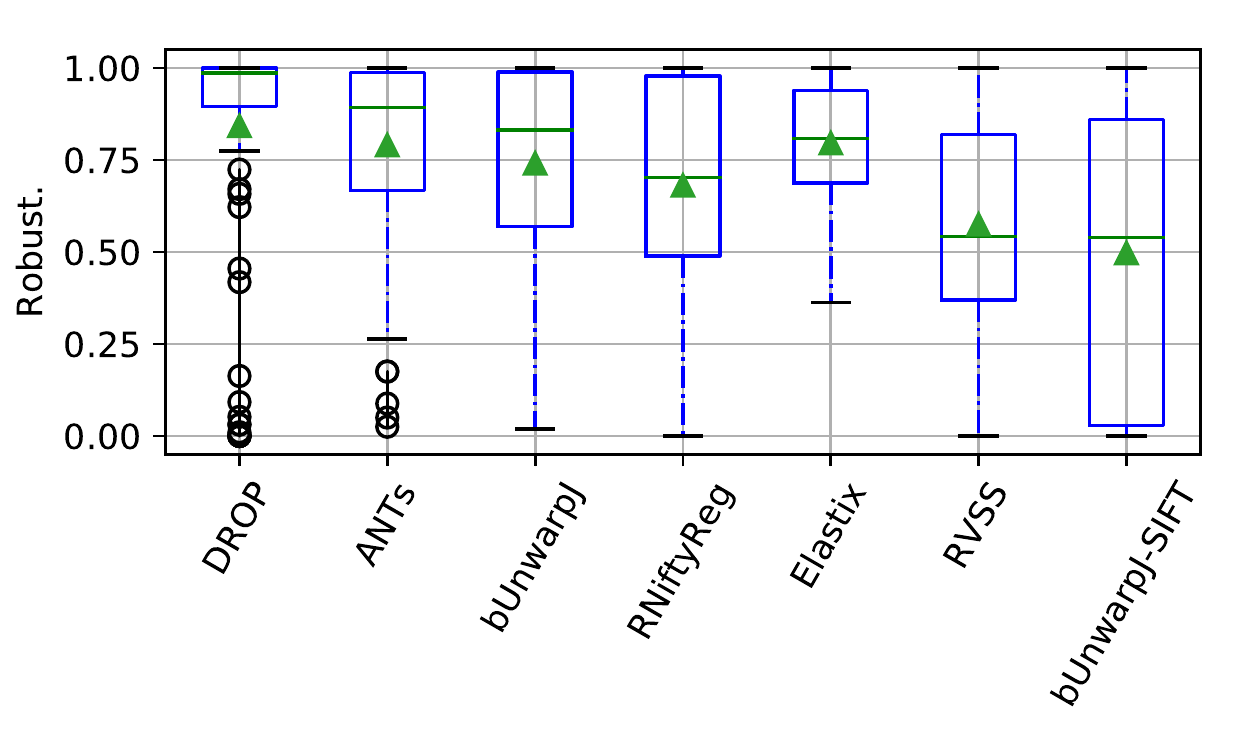}
    \includegraphics[width=0.49\textwidth]{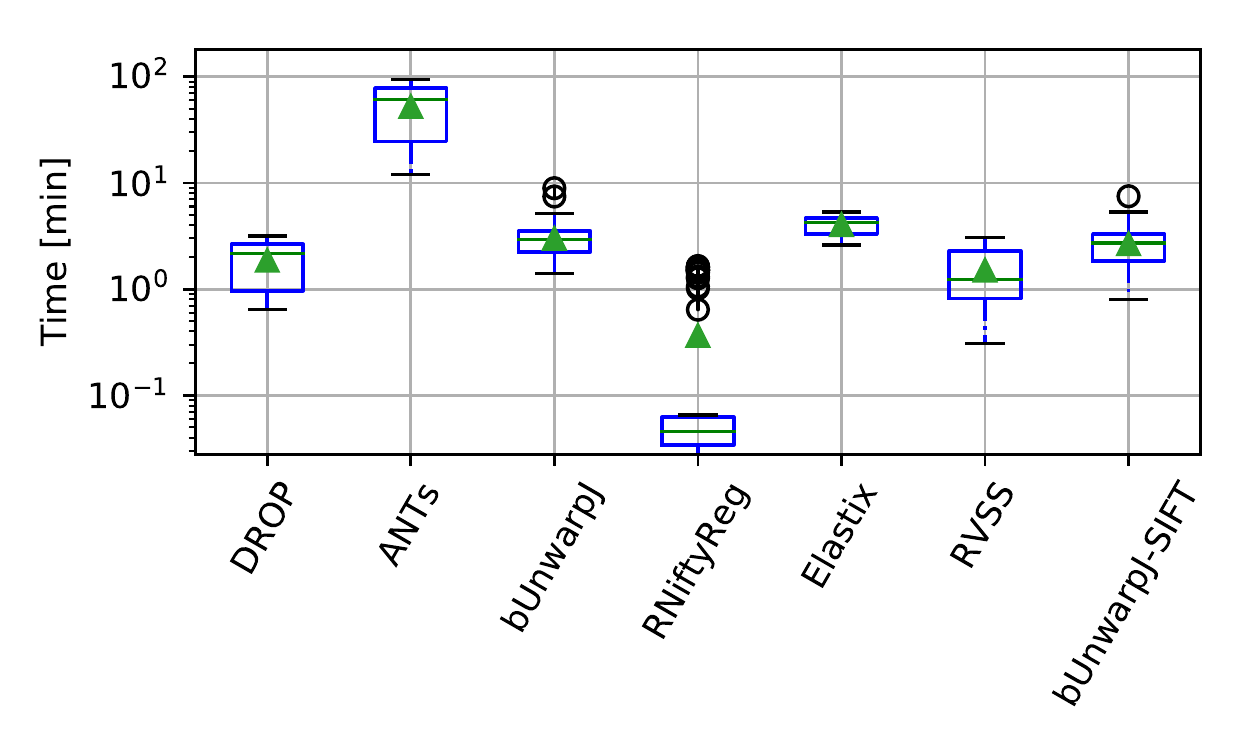}
  \caption{\label{fig:errorbar}Box-plot visualisation of particular measure distribution on dataset scope $10k$: (top-left) median rTRE, (top-right) maximal rTRE, (bottom-left) robustness and (bottom-right) execution time.}
\end{figure}

\begin{figure}[t]
  \centering
    \includegraphics[width=0.49\textwidth]{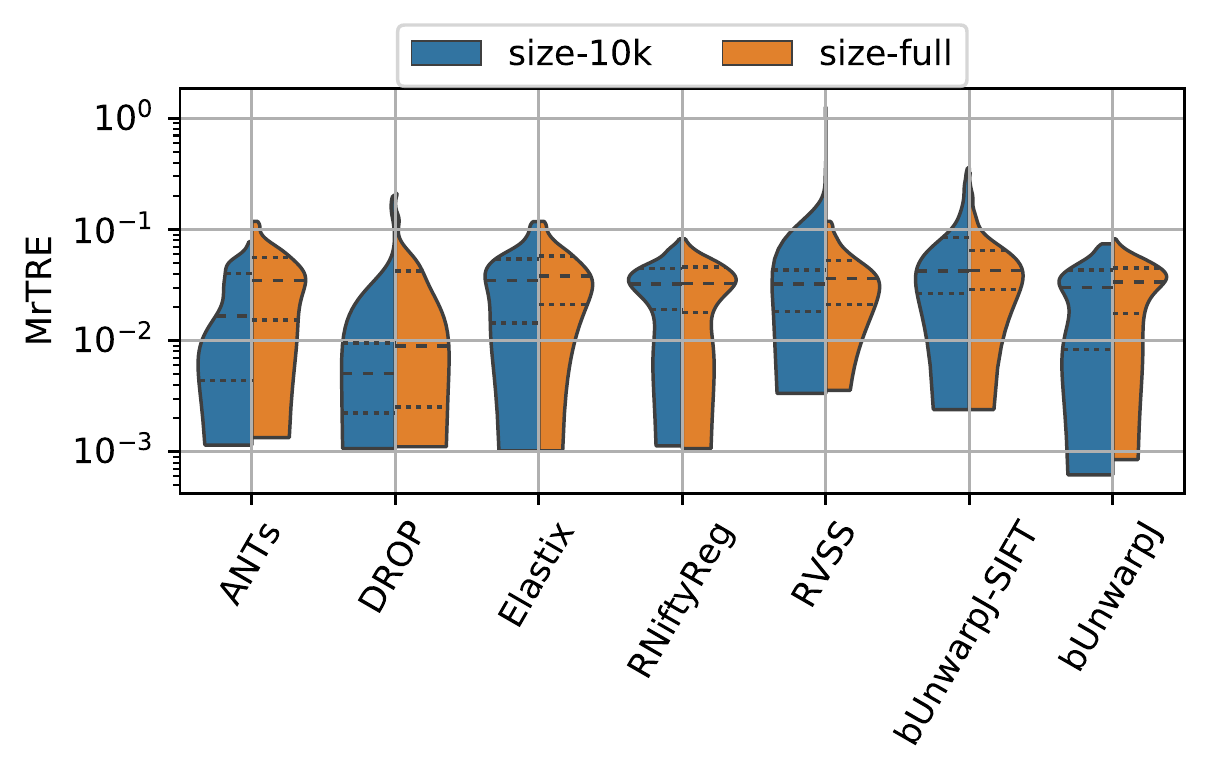}
    \includegraphics[width=0.49\textwidth]{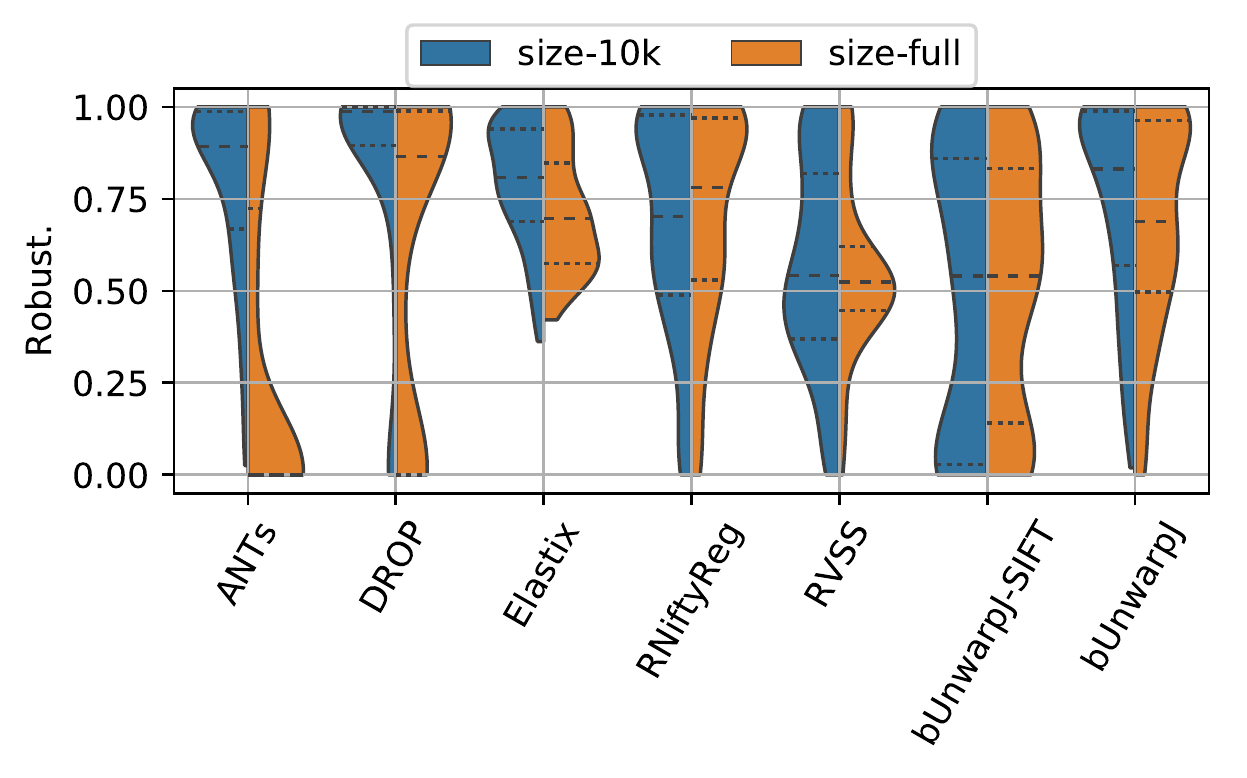}
  \caption{\label{fig:distrib}Distribution comparison between the two dataset scopes - $10k$ and $full$ size, presented on (left) median rTRE and (right) robustness.}
\end{figure}

We have observed a gap between results on $10k$ and $full$ dataset scope since some methods are not able to work on some very large images from $full$ scope, typically they fail on pixel indexing or saving larger images then 32k pixels.
The summary results on both dataset scopes are presented in Tab.~\ref{tab:results}.

For the rather technical/implementation limitations, we present the most visual comparison on the $10k$ dataset scope, except comparison of the two scopes in Fig.~\ref{fig:distrib}.


\begin{figure}[t]
  \centering
    \includegraphics[width=0.49\textwidth,height=0.2\textheight,keepaspectratio]{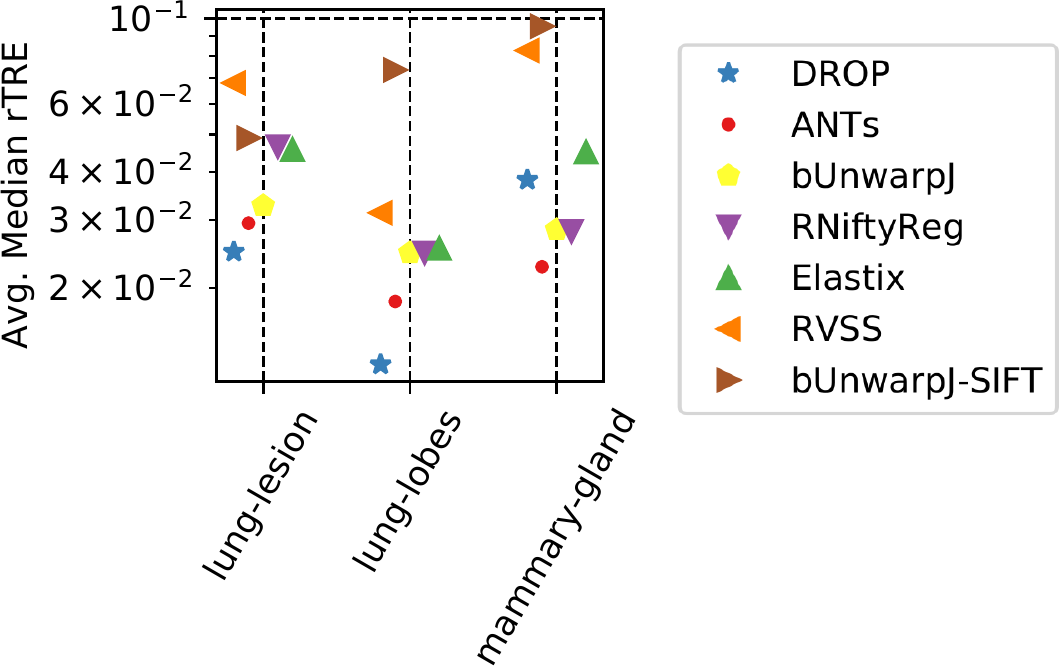}
    \hskip5mm
    \includegraphics[width=0.49\textwidth,height=0.2\textheight,keepaspectratio]{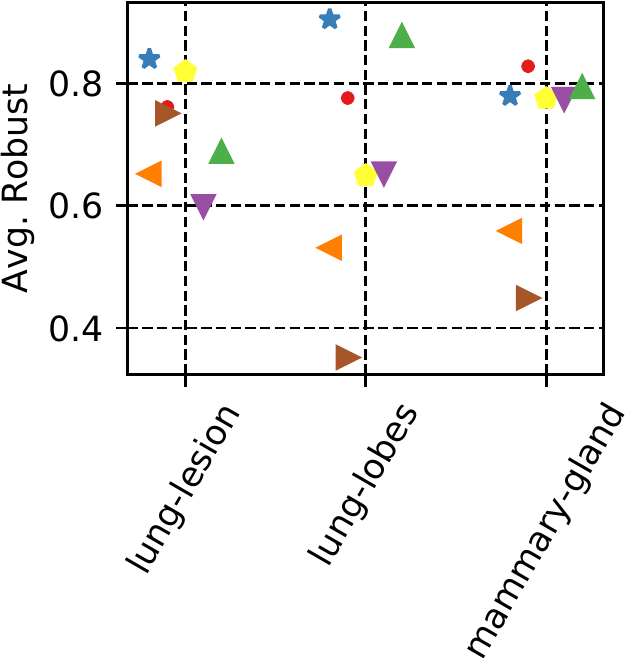}
  \caption{\label{fig:tissue}Performance visualisation of particular methods depending on tissue kind: (left) average median rTRE and (right) robustness on dataset scope $10k$.}
\end{figure}

First, we show the radar chart in Fig.\ref{fig:radar} aggregating over all major metrics.
We can see that most of the performance curves are parallel, meaning that a particular method usually performs better than its competitors in all aspects.

To show detail inside how the integrated methods perform over the CIMA datasets in the four main metrics: median and max rTRE per image (denoted as MrTRE and SrTRE respectively), robustness and execution time; the methods are ordered from left to right by increasing $AMrTRE$.
In the Fig.~\ref{fig:errorbar}, we can also observe a correlation among all quality measures (MrTRE, SrTRE and Robust.) for all methods, meaning that DROP has lowest AMrTRE and also highest robustness.

Fig.~\ref{fig:distrib} addresses the comparison between the two dataset scopes.
Looking at the robustness, we can see a significant increase of completely failed registrations for ANTs and DROP on while transiting from $10k$ to $full$ scope compare to other methods where we can see a less significant quality drop.
The MrTRE increase consistently for all methods while transiting from $10k$ to $full$ scope.

We have also examined performances depending on the tissue types as they significantly differ in appearance - repetitive texture patterns and tissue separability from the background (and size, in particular, for $full$ image scope), see Fig.~\ref{fig:tissue}.
In this perspective, there is no clear trend for all method's performances across tissues, for example, Elastix and DROP perform better on lung lobes compare to other methods.




\section{Conclusion}

In this report, we briefly introduced the developed image registration framework using dataset with landmark annotations and presented the main application use-cases.
We described the CIMA histology dataset with some more information about image sensing, landmarks annotation and image pairing.
Later we presented selected standard image registration methods integrated into BIRL framework and their results on the presented CIMA dataset with illustrative visualisations. 

Hence, any future work uses this as a starting point and can be compared with these result as a baseline.


\bibliographystyle{unsrt}
\bibliography{references}

\begin{thebibliography}{10}

\bibitem{Maintz1998}
J.B.~B Maintz and M.A.~A Viergever.
\newblock {A survey of medical image registration}.
\newblock {\em Medical image analysis}, 2(1):1--36, mar 1998.

\bibitem{zitova2003}
Barbara Zitov{\'{a}}, Jan Flusser, and B.~Zitova.
\newblock {Image registration methods: a survey}.
\newblock {\em Image and vision computing}, 21(11):977--1000, 2003.

\bibitem{Salvi2007}
Joaquim Salvi, Carles Matabosch, David Fofi, and Josep Forest.
\newblock {A review of recent range image registration methods with accuracy
  evaluation}.
\newblock {\em Image and Vision Computing}, 25(5):578--596, 2007.

\bibitem{Sotiras2013}
Aristeidis Sotiras, Christos Davatzikos, and Nikos Paragios.
\newblock {Deformable Medical Image Registration: A Survey}.
\newblock {\em Trans Med Imaging}, 32(7):1153--1190, jul 2013.

\bibitem{Alam2017}
Fakhre Alam, Sami~Ur Rahman, Muhammad Hassan, and Adnan Khalil.
\newblock {An investigation towards issues and challenges in medical image
  registration}.
\newblock {\em Journal of Postgraduate Medical Institute}, 31(3):224--233,
  2017.

\bibitem{Haskins2019}
Grant Haskins, Uwe Kruger, and Pingkun Yan.
\newblock {Deep Learning in Medical Image Registration: A Survey}.
\newblock 2019.

\bibitem{Alam2019}
Fakhre Alam and Sami~Ur Rahman.
\newblock {Challenges and Solutions in Multimodal Medical Image Subregion
  Detection and Registration}.
\newblock {\em Journal of Medical Imaging and Radiation Sciences},
  50(1):24--30, 2019.

\bibitem{Deniz2015}
Oscar D{\'{e}}niz, David Toomey, Catherine Conway, Gloria Bueno, Al., Catherine
  Conway, and Gloria Bueno.
\newblock {Multi-stained whole slide image alignment in digital pathology}.
\newblock In {\em Proc. SPIE Medical Imaging}, volume 9420, page 94200Z, 2015.

\bibitem{Gurcan2009}
M~Gurcan, L~E Boucheron, A~Can, A~Madabhushi, N~M Rajpoot, and B~Yener.
\newblock {Histopathological image analysis: A review}.
\newblock {\em IEEE Reviews in Biomedical Engineering}, 2:147--171, 2009.

\bibitem{West1997}
J~West, JM~M Fitzpatrick, and MY~Y Wang.
\newblock {Comparison and evaluation of retrospective intermodality brain image
  registration techniques}.
\newblock {\em Journal of Computer Assisted Tomography}, 21(4):554--568, 1997.

\bibitem{Murphy2011}
Keelin Murphy, Bram {Van Ginneken}, Joseph~M. Reinhardt, Al., Et~Al., Sven
  Kabus, Kai Ding, Xiang Deng, Kunlin Cao, Kaifang Du, Gary~E. Christensen,
  Vincent Garcia, Tom Vercauteren, Nicholas Ayache, Olivier Commowick, Grgoire
  Malandain, Ben Glocker, Nikos Paragios, Nassir Navab, Vladlena Gorbunova, Jon
  Sporring, Marleen {De Bruijne}, Xiao Han, Mattias~P. Heinrich, Julia~a.
  Schnabel, Mark Jenkinson, Cristian Lorenz, Marc Modat, Jamie~R. McClelland,
  Sebastien Ourselin, Sascha E~a Muenzing, Max~a. Viergever, Dante {De Nigris},
  D.~Louis Collins, Tal Arbel, Marta Peroni, Rui Li, Gregory~C. Sharp,
  Alexander Schmidt-Richberg, Jan Ehrhardt, Ren{\'{e}} Werner, Dirk Smeets,
  Dirk Loeckx, Gang Song, Nicholas Tustison, Brian Avants, James~C. Gee, Marius
  Staring, Stefan Klein, Berend~C. Stoel, Martin Urschler, Manuel Werlberger,
  Jef Vandemeulebroucke, Simon Rit, David Sarrut, and Josien P~W Pluim.
\newblock {Evaluation of registration methods on thoracic CT: The EMPIRE10
  challenge}.
\newblock {\em IEEE Transactions on Medical Imaging}, 30(11):1901--1920, 2011.

\bibitem{Metzger2012}
Gregory~J. Metzger, Stephen~C. Dankbar, Jonathan Henriksen, Al., Anthony~E.
  Rizzardi, Nikolaus~K. Rosener, and Stephen~C Schmechel.
\newblock {Development of multigene expression signature maps at the protein
  level from digitized immunohistochemistry slides.}
\newblock {\em PloS One}, 7(3):e33520, jan 2012.

\bibitem{Garcia2010}
Idoia Garcia, Gemma Mayol, Eva Rodr{\'{i}}guez, Mariona Su{\~{n}}ol, Timothy~R.
  Gershon, Jos{\'{e}} R{\'{i}}os, Nai Kong~V. Cheung, Mark~W. Kieran, Rani~E.
  George, Antonio~R. Perez-Atayde, Carla Casala, Patricia Galv{\'{a}}n, Carmen
  de~Torres, Jaume Mora, and Cinzia Lavarino.
\newblock {Expression of the neuron-specific protein CHD5 is an independent
  marker of outcome in neuroblastoma}.
\newblock {\em Molecular Cancer}, 9:1--14, 2010.

\bibitem{Novakovic2012}
Zana~Saratlija Novakovic, Merica~Glavina Durdov, Livia Puljak, Marijan Saraga,
  Dragan Ljutic, Tomislav Filipovic, Zvonimir Pastar, Antonia Bendic, and
  Katarina Vukojevic.
\newblock {The interstitial expression of alpha-smooth muscle actin in
  glomerulonephritis is associated with renal function}.
\newblock {\em Medical Science Monitor}, 18(4):235--240, 2012.

\bibitem{Thul2018}
Peter~J. Thul and Cecilia Lindskog.
\newblock {The human protein atlas: A spatial map of the human proteome}.
\newblock {\em Protein Science}, 27(1):233--244, 2018.

\bibitem{Chin2007}
Mark~H. Chin, Alex~B. Geng, Arshad~H. Khan, Wei~Jun Qian, Vladislav~A. Petyuk,
  Jyl Boline, Shawn Levy, Arthur~W. Toga, Richard~D. Smith, Richard~M. Leahy,
  and Desmond~J. Smith.
\newblock {A genome-scale map of expression for a mouse brain section obtained
  using voxelation}.
\newblock {\em Physiological Genomics}, 30(3):313--321, 2007.

\bibitem{Lopez2014}
Xavier~Moles Lopez, Paul Barbot, Yves~R{\'{e}}mi {Van Eycke}, Laurine Verset,
  Anne~Laure Tr{\'{e}}pant, Lionel Larbanoix, Isabelle Salmon, and Christine
  Decaestecker.
\newblock {Registration of whole immunohistochemical slide images: An efficient
  way to characterize biomarker colocalization}.
\newblock {\em Journal of the American Medical Informatics Association},
  22(1):86--99, 2014.

\bibitem{Song2013}
Yi~Song, Darren Treanor, AndrewJ Bulpitt, Al., and Derek~R. Magee.
\newblock {3D reconstruction of multiple stained histology images}.
\newblock {\em Journal of Pathology Informatics}, 4(2):7, 2013.

\bibitem{Kartasalo2016}
Kimmo Kartasalo, Leena Latonen, Tapio Visakorpi, P~Nykter, and Pekka
  Ruusuvuori.
\newblock {Benchmarking of image registration methods for 3D tissue
  reconstruction.}
\newblock In {\em IEEE International Conference on Image Processing,}, number
  269474, pages 2--6, 2016.

\bibitem{Kartasalo2018}
Kimmo Kartasalo, Leena Latonen, Tapio Visakorpi, Matti Nykter, and Pekka
  Ruusuvuori.
\newblock {Comparative Analysis of Tissue Reconstruction Algorithms for 3D
  Histology.}
\newblock {\em Bioinformatics}, 34(17):2360--2364, 2018.

\bibitem{Borovec2013}
Jiri Borovec, Jan Kybic, Michal Bu{\v{s}}ta, Carlos Ortiz-de Solorzano, and
  Arrate Munoz-Barrutia.
\newblock {Registration of multiple stained histological sections}.
\newblock In {\em IEEE International Symposium on Biomedical Imaging (ISBI)},
  pages 1034--1037, San Francisco, 2013.

\bibitem{Kybic2014}
Jan Kybic and Jiri Borovec.
\newblock {Automatic simultaneous segmentation and fast registration of
  histological images}.
\newblock In {\em IEEE International Symposium on Biomedical Imaging (ISBI)},
  pages 774 -- 777, 2014.

\bibitem{Borovec2018}
Jiri Borovec, Arrate Munoz-Barrutia, and Jan Kybic.
\newblock {Benchmarking of Image Registration Methods for Differently Stained
  Histological Slides}.
\newblock In {\em IEEE International Conference on Image Processing (ICIP)},
  pages 3368--3372, Athens, 2018.

\bibitem{Castillo2009}
Richard Castillo, Edward Castillo, Rudy Guerra, Valen~E Johnson, Travis
  McPhail, Amit~K Garg, Thomas Guerrero, and Al.
\newblock {A framework for evaluation of deformable image registration spatial
  accuracy using large landmark point sets.}
\newblock {\em Physics in medicine and biology}, 54(7):1849--1870, 2009.

\bibitem{Ou-TMI2014}
Y~Ou, H~Akbari, M~Bilello, X~Da, and C~Davatzikos.
\newblock {Comparative Evaluation of Registration Algorithms in Different Brain
  Databases With Varying Difficulty: Results and Insights}.
\newblock {\em IEEE Transactions Med. Imag.}, 33(10):2039--2065, 2014.

\bibitem{Christensen2006}
Gary~E Christensen, Xiujuan Geng, Jon~G Kuhl, Joel Bruss, Thomas~J Grabowski,
  Imran~a Pirwani, Michael~W Vannier, John~S Allen, and Hanna Damasio.
\newblock {Introduction to the Non-Rigid Image Registration Evaluation
  Project}.
\newblock In {\em Lecture Notes in Computer Science}, volume 4057, pages
  128--135, 2006.

\bibitem{Klein2009}
Arno Klein, Jesper Andersson, Babak a.~B.A. Ardekani, John Ashburner, Brian
  Avants, Ming~Chang Chiang, Gary~E. Christensen, D.~Louis Collins, James Gee,
  Pierre Hellier, Joo~Hyun Song, Mark Jenkinson, Claude Lepage, Daniel
  Rueckert, Paul Thompson, Tom Vercauteren, Roger~P. Woods, J.~John Mann,
  Ramin~V. Parsey, Al., and Et~Al.
\newblock {Evaluation of 14 nonlinear deformation algorithms applied to human
  brain MRI registration}.
\newblock {\em Neuroimage}, 46(3):786--802, 2009.

\bibitem{Marstal2019}
K.~Marstal, F.~Berendsen, N.~Dekker, M.~Staring, and S.~Klein.
\newblock {The continuous registration challenge: Evaluation-as-a-service for
  medical image registration algorithms}.
\newblock In {\em Proceedings - International Symposium on Biomedical Imaging},
  volume 2019-April, pages 1399--1402. IEEE, 2019.

\bibitem{Brock-IJRO2010}
Kristy~K Brock.
\newblock {Results of a Multi-Institution Deformable Registration Accuracy
  Study (MIDRAS)}.
\newblock {\em International Journal of Radiation Oncology, Biology, Physics},
  76(2):583--596, 2010.

\bibitem{Avants2008}
B.~B. Avants, C.~L. Epstein, M.~Grossman, and J.~C. Gee.
\newblock {Symmetric diffeomorphic image registration with cross-correlation:
  Evaluating automated labeling of elderly and neurodegenerative brain}.
\newblock {\em Medical Image Analysis}, 12(1):26--41, 2008.

\bibitem{Arganda2006}
Ignacio Arganda-Carreras, C.~Sorzano, Roberto Marabini, Al., Mar{\'{i}}a
  {Jos{\'{e}} Carazo}, Carlos Ortiz-de Solorzano, and Jan Kybic.
\newblock {Consistent and elastic registration of histological sections using
  vector-spline regularization}.
\newblock In {\em Computer Vision Approaches to Medical Image Analysis}, volume
  4241, pages 85----95, 2006.

\bibitem{FIJI2012}
J~Schindelin, I~Arganda-Carreras, E~Frise, and Others.
\newblock {Fiji: an open-source platform for biological-image analysis}.
\newblock {\em Nature Methods}, 9(7):676--682, 2012.

\bibitem{Glocker2008}
B.~Glocker, N.~Komodakis, G.~Tziritas, and N.~Navab.
\newblock {Dense image registration through MRFs and efficient linear
  programming}.
\newblock {\em Medical Image Analysis}, 12(6):731--741, 2008.

\bibitem{Glocker2011}
Ben Glocker, Aristeidis Sotiras, Nikos Komodakis, Nikos Paragios, and Al.
\newblock {Deformable Medical Image Registration: Setting the State of the Art
  with Discrete Methods}.
\newblock {\em Annual Review of Biomedical Engineering}, 13(1):219--244, 2011.

\bibitem{Klein2010}
S.~Klein, M.~Staring, and K.~Murphy.
\newblock {Elastix: a toolbox for intensity-based medical image registration}.
\newblock {\em Medical Imaging, IEEE}, 29(1), 2010.

\bibitem{Ourselin2001}
S.~Ourselin, A.~Roche, and G.~Subsol.
\newblock {Reconstructing a 3D structure from serial histological sections}.
\newblock {\em Image and Vision Computing}, 19(1-2):25--31, 2001.

\bibitem{Modat2014}
Marc Modat, David Cash, Pankaj Daga, Gavin Winston, John S.~Duncan, and
  Sébastien Ourselin.
\newblock A symmetric block-matching framework for global registration.
\newblock volume 9034, page 90341D, 03 2014.

\bibitem{lowe04}
D.~Lowe.
\newblock Distinctive image features from scale-invariant keypoints.
\newblock 60(2):91--110, 2004.

\bibitem{Fernandez2002}
R.~Fernandez-Gonzalez, A.~Jones, E.~Garcia-Rodriguez, P.Y. Chen, A~Idica, S.J.
  Lockett, M.H. Barcellos-Hoff, and C.~Ortiz~de Sol{\'o}rzano.
\newblock System for combined three-dimensional morphological and molecular
  analysis of thick tissue specimens.
\newblock {\em Microscopy Research \& Techniques}, (59):522--530, 2002.

\end{thebibliography}

\end{document}